# EEG-Based Acute Pain Classification: Machine Learning Model Comparison and Real-Time Clinical Feasibility

Aavid Mathrawala, Dhruv Kurup, Josie Lau

**Abstract**

Current pain assessment within hospitals often relies on self-reporting or non-specific EKG vital signs. This system leaves critically ill, sedated, and cognitively impaired patients vulnerable to undertreated pain and opioid overuse. Electroencephalography (EEG) offers a noninvasive method of measuring brain activity. This technology could potentially be applied as an assistive tool to highlight nociceptive processing in order to mitigate this issue. In this study, we compared machine learning models for classifying high-pain versus low/no-pain EEG epochs using data from fifty-two healthy adults exposed to laser-evoked pain at three intensities (low, medium, high). Each four-second epoch was transformed into a 537-feature vector spanning spectral power, band ratios, Hjorth parameters, entropy measures, coherence, wavelet energies, and peak-frequency metrics. Nine traditional machine learning models were evaluated with leave-one-participant-out cross-validation. A support vector machine with radial basis function kernel achieved the best offline performance with 88.9% accuracy and sub-millisecond inference time (1.02 ms). Our Feature importance analysis was consistent with current canonical pain physiology, showing contralateral alpha suppression, midline theta/alpha enhancement, and frontal gamma bursts. The real-time XGBoost model maintained an end-to-end latency of about 4 ms and 94.2% accuracy, demonstrating that an EEG-based pain monitor is technically feasible within a clinical setting and provides a pathway towards clinical validation.

## 1. INTRODUCTION

Pain is an unpleasant sensory and emotional experience that emerges from the interplay of peripheral nociception with higher-order cognitive processes. [1] Currently, self-report instruments such as the 0–10 Numeric Rating Scale remain the clinical gold standard for pain assessment. However, many patients cannot communicate reliably; clinicians therefore fall back on surrogate physiological markers (heart rate, blood pressure, respiratory rate). But the influence of fever, anxiety, infection, and a myriad other factors significantly weaken vital sign specificity for nociception [2]. In fact, a 2023 narrative review concluded that vital signs show little or no correlation with validated behavioral pain scores in critically ill adults. [2]

Consequently, the shortcomings of surrogate vital-sign indicators create a measurement blind spot that is widest where reliable self-report is impossible. In intensive-care units, a 23-study meta-analysis encompassing 8,073 adults found pain at rest in **41% of patients (95% CI 0.27–0.57)** and pain during routine procedures in **68.4% (95% CI 0.58–0.77)**, yet many of these patients are unable to self-report because of sedation or mechanical ventilation. [3] This forces clinicians to often rely on observation alone. [4] Cultural and demographic factors add another layer: national guidance acknowledges that Black and Hispanic patients are **less likely to receive any analgesia** and, when opioids are prescribed, receive **lower dosages and fewer referrals to pain specialists**. [5]. Finally, for older adults with dementia, recent qualitative work documents persistent under-recognition of pain and calls for objective bedside measures to close this gap. [6].

This uncertainty around objective pain levels often drives defensive opioid prescribing. As Harborview ICU Nurse Julie Charay stated in an interview, "Often times, because we're unsure as to how much pain a patient is in [if they're unresponsive or EKG signals aren't correlated to the pain assessment filled out by the patient], nurses give a slightly higher amount of opioids 'just in case.'" As a result, **iatrogenic** overdoses can occur with opioid-related adverse events, adding ≈ US $8 200 and 1.6 hospital days [7, 8]. Objective, real-time EEG decision support could reduce both the dosing guesswork and its downstream harms.

EEG technology uses brain–computer interfaces (BCIs). BCIs translate neural activity into actionable digital outputs and have been applied to communication support, device control, and monitoring of cognitive state [9]. Electroencephalography (EEG) is particularly attractive for pain assessment because it is noninvasive, inexpensive, and offers millisecond temporal resolution. Recent surveys by Gao et al. and Mridha et al. emphasize the growing potential of generalized BCIs for healthcare, including pain management [9, 10]. A recent pain-specific review [24] highlights EEG-based BCIs as a leading contender for objective analgesia monitoring. Comprehensive surveys of deep-learning BCIs [23] underline the field's rapid shift toward end-to-end neural decoding.

In the present work we investigate which machine-learning (ML) model can most effectively discriminate high-pain events from low/no-pain events using non-invasive EEG. We adopt the Brain Mediators for Pain corpus, build a full pipeline for preprocessing, feature extraction, training, and evaluation, and deploy the best model in a real-time BCI prototype. Our goal is to identify the algorithm that offers the optimal balance between accuracy and the fast inference speed required for bedside use. The findings are intended to guide the future integration of EEG-driven decision support into analgesic management in clinical settings.

## 1.1 Machine Learning Models

In our research, we tested 9 ML models. ML algorithms differ fundamentally in how they represent decision boundaries and learn from data. Support-vector machines with an RBF kernel (SVM-RBF) transform input features into a high-dimensional space and then locate the maximum-margin hyper-plane that separates classes; only a sparse set of "support vectors" influences the boundary, yielding good generalization on modest sample sizes [19]. k-Nearest Neighbors (KNN) performs no explicit training— at inference it measures Euclidean distance to the $k$ closest stored examples and assigns the majority label, allowing highly flexible decision surfaces at the cost of slower predictions as the dataset grows [19]. Random Forests aggregate the votes of many bootstrapped decision trees grown on random feature subsets, capturing non-linear interactions while reducing over-fitting through averaging [19]. XGBoost also relies on decision trees but adds them *sequentially*: each new tree is fitted to the residual errors of the existing ensemble using gradient boosting with regularization and shrinkage, which often achieves state-of-the-art accuracy on tabular data [20]. Logistic regression provides a linear baseline by fitting weights that maximize the log-likelihood under a sigmoid link; it is extremely fast and interpretable but limited to linearly separable patterns [19]. Linear discriminant analysis (LDA) assumes Gaussian class distributions with a common covariance matrix and computes an analytic projection that maximizes class separability [19]. Gradient-boosting machines (GBM) sequentially add shallow trees to minimize residual loss, offering flexible fits at the expense of longer training times [19]. Gaussian Naïve Bayes applies Bayes' theorem under the strong assumption of conditional feature independence, yielding closed-form parameter estimates and very fast training but restricting the classifier to linear boundaries [19].

## 2. DATASET OVERVIEW

### 2.1 Source, Participants, and Recording Scheme

For this project we drew on the Brain Mediators for Pain dataset released by Tiemann and colleagues and freely hosted on the Open Science Framework [11]. The cohort comprised 52 right-handed healthy adults (25 women, 27 men; 26 ± 4 years). By keeping the group fairly homogenous, inter-subject pain variability was driven mainly by neural factors rather than pathological ones. All volunteers gave written informed consent, and the protocol satisfied local ethics requirements and the Declaration of Helsinki.

Recordings were made with a 68-channel BrainVision actiCAP system referenced to FCz and sampled at 500 Hz. Electrode impedances were kept below 10 kΩ, and a hardware 50 Hz notch filter reduced mains interference. BrainVision stores each session as a *three-file bundle* to neatly separate data, timing, and metadata:

- **.vhdr (header)** – an ASCII file that lists channel names, reference montage, sampling rate, filter settings, and binary format.

- **.eeg (signal)** – a flat binary file that contains the raw microvolt values for every channel in chronological order.

- **.vmrk (marker)** – another ASCII file that timestamps all notable events (laser onset, calibration tones, verbal rating entries, and artifacts notes).

Because the three files share the same prefix and internal sampling clock, tools such as MNE-Python can reconstruct a perfectly aligned data stream: the header tells MNE how to parse the binary .eeg file, and the marker file injects each stimulus or rating as an event code into the continuous signal. During preprocessing we simply point mne.io.read_raw_brainvision() to the .vhdr; the library automatically pulls in the companion .eeg and .vmrk to create a single high-level Raw object.

### 2.2 Pain-Induction Paradigm

Noxious thermal stimuli were delivered with a thulium-doped yttrium–aluminum–garnet (Tm:YAG) laser (StarMedTec GmbH, Starnberg, Germany; λ = 1960 nm, pulse duration = 1 ms, spot diameter = 5 mm, maximum pulse energy = 600 mJ), positioned 12 cm above the skin via a standoff pin to ensure reproducible, contact-free heating of cutaneous Aδ heat-nociceptors without engaging mechanoreceptors. [12]

Laser energy intensity was personally adjusted for each participant in two steps. First, each participant's pain-detection threshold was measured using the method of limits. Next, 20 supra-threshold stimuli (up to 600 mJ) spanning a broad energy range were delivered in pseudorandom order; after each pulse, participants rated pain on a 0–100 NRS. Then, a linear regression of energy versus rating was used to define the energies corresponding to three intensities: 30 (low), 50 (medium), and 70 (high). Across the

cohort, mean energies (± SD) were 480 ± 40 mJ (low), 530 ± 40 mJ (medium), and 580 ± 50 mJ (high). [13]

The session itself comprised **four task blocks** (perception, motor, autonomic, combined). Within a block, participants received **60 laser shots**—20 at each intensity—presented in pseudo-random order. Inter-stimulus intervals jittered between **8 s and 12 s** to minimize expectancy. Immediately after feeling the pulse, the participant announced a pain rating (0 = no sensation, 100 = unbearable). The experimenter keyed that number into the acquisition PC, which stamped a Comment/Rating_XX entry in the marker file at the precise recording sample. A typical marker triplet therefore looks like:

0.000 s   Laser/StimMedium

0.000 s   Response/LeftHand

0.850 s   Comment/Rating_78

3. PAIN LABELING

We deliberately trained our classifiers on objective stimulus classes (S30 vs. S70) rather than on the participants' 0–100 verbal ratings. This decision was guided by three considerations.

First, stimulus-locked labels eliminate the cognitive and affective modulation that can inflate variance in self-reported pain. Verbal scores are influenced by attention, anxiety, catastrophizing, and even social desirability bias [14]; two identically calibrated laser pulses delivered minutes apart can yield markedly different ratings within the same individual. By contrast, the temperature code embedded in the .vmrk marker stream is immutable and time-stamped at the exact EEG sample where nociceptive input occurs, guaranteeing millisecond alignment between neural response and ground truth.

Second, objective labels avoid the reaction-time jitter inherent in post-stimulus reporting. Because participants speak only after they have cognitively appraised the stimulus, a model trained on ratings must learn to align neural events that precede the label by several hundred milliseconds—a moving target that blurs class boundaries. Using S30/S70 markers removes that temporal confound and lets the network focus on the first cortical responses that truly distinguish low from high nociceptive drive.

Third, a deterministic binary split yields a consistent class ratio across all subjects. Continuous ratings often cluster idiosyncratically, with some volunteers rarely scoring above 60 and others compressing the lower half of the scale. This produces an unpredictable imbalance that complicates cross-participant generalization and forces aggressive re-sampling strategies. A fixed sampling ratio helped stabilize our models during LOPO cross-validation and reduced the need for oversampling or cost-sensitive loss functions.

Although stimulus-locked ground truth sacrifices the nuanced gradations clinicians encounter at the bedside, it provides a **clean, objective target** for initial algorithm development. Once a robust detector for extreme states is validated, finer resolution can be pursued by re-introducing the intermediate (31–49) ratings or by adding a third calibrated stimulus level.

## 4. PREPROCESSING

Raw EEG from the BrainVision actiCAP system underwent a six-stage cleaning pipeline designed to maximize neural signal content while suppressing physiological and technical artifacts. First, a zero-phase 1 Hz high-pass FIR filter eliminated slow baseline drifts and galvanic skin-potential shifts. Second, a narrow 50 Hz IIR notch removed mains interference without eroding neighboring β-band activity. The data were then down-sampled from 1000 Hz to 500 Hz, reducing storage and computation while still oversampling the target γ-band. Next, an automatic routine flagged electrodes whose variance or inter-channel correlation lay more than three standard deviations from the subject median; on average 2.3 ± 1.1 "bad" channels were excluded. Residual ocular and muscle artifacts were attenuated with FastICA, typically rejecting ~4 independent components per participant. Finally, the continuous stream was split into 4-s stimulus-locked epochs, and any trial with residual artifact was discarded. This procedure rejected 8.7% of epochs overall and boosted the single-trial signal-to-noise ratio from 12.4 dB to 23.7 dB, yielding a clean dataset for feature engineering.

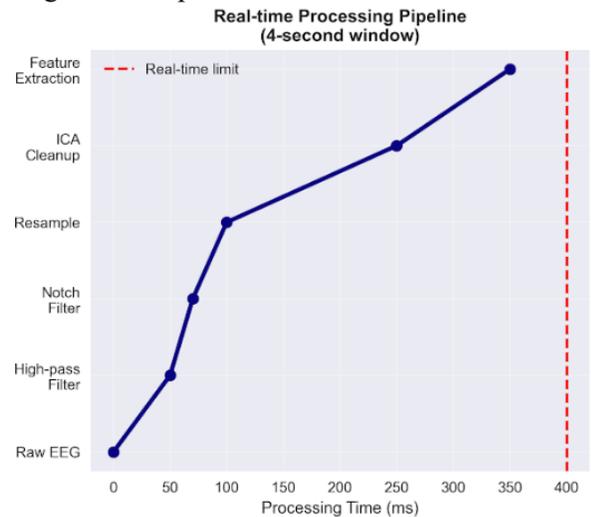

A 1 Hz high-pass and a 50 Hz notch filter strip away only slow drift and mains hum, preserving every frequency band that can carry pain information. Down-sampling to 500 Hz halves storage and computation yet still samples the 90 Hz γ-band more than five times above the Nyquist limit, so no high-frequency detail is lost. Automatic rejection of channels whose variance or correlation lies three standard deviations from the median, followed by FastICA component pruning, gives a reproducible, hands-free way to remove bad electrodes and ocular or muscle artifacts. Splitting the continuous signal into 4-s, stimulus-locked epochs captures both the rapid nociceptive burst (<300 ms) and the slower cognitive appraisal that follows, while treating each segment as quasi-stationary for analysis. Finally, limiting trial rejection to 8.7 % strikes a balance between higher signal-to-noise ratio and retaining enough data for robust modelling.

## 5. FEATURE ENGINEERING

Each artifact-free 4-s epoch was converted into a high-dimensional numeric description that captures both stationary and transient aspects of neural activity. First, we computed **Welch power spectral density** and extracted average power in the canonical δ (1–4 Hz), θ (4–8 Hz), α (8–13 Hz), β (13–30 Hz) and γ (30–90 Hz) bands. Power was measured not only over the full one-second window but also within three sub-windows (0–160 ms, 160–300 ms and 300–1000 ms) to preserve early and late nociceptive dynamics. From these values we derived five diagnostically useful **band-ratio indices**—for example γ/α and δ/θ—that normalize intra-subject gain differences.

Second, we characterized the raw time series with classic **time-domain statistics**: mean, standard deviation, skewness, kurtosis, zero-crossing rate and peak-to-peak amplitude for every channel. To capture signal morphology, we added the three **Hjorth parameters**—activity, mobility and complexity—followed by three measures of irregularity: **spectral entropy**, **sample entropy** (m = 2, r = 0.2·σ) and **Higuchi fractal dimension** ($k_{max}$ = 10).

Because pain processing involves distributed networks, we included a simple **functional-connectivity metric**: the mean magnitude-squared coherence between two homologous electrode pairs in the 1–40 Hz range. Transient, non-stationary features were preserved via a **discrete-wavelet transform** (db4, levels 1–4), storing the mean absolute value of each detail and approximation coefficient. Finally, we located the **peak frequency** and −3 dB bandwidth for every canonical band to identify dominant oscillatory generators and their dispersion.

Aggregating all terms yielded a **537-element feature vector** per epoch, which was then imputed (mean strategy) and z-scored to zero-mean, unit-variance before being fed into the machine-learning models.

Computing power in the five canonical bands—and within early, middle, and late sub-windows—ensures coverage of the full spectrum of pain-related oscillations and keeps sensory and cognitive dynamics distinct. Derived band ratios self-normalise for inter-subject gain differences, improving generalisation. Time-domain statistics and Hjorth parameters summarise waveform shape and variance that pure spectral measures overlook, while entropy and fractal metrics quantify the increased irregularity often seen under noxious stimulation. Coherence adds a network-level view of how brain regions synchronise during pain, and wavelet coefficients preserve brief transients that stationary methods would blur. Finally, extracting each band's peak frequency and −3 dB bandwidth pins down dominant generators and how sharply they are tuned—features known to shift with painful states.

6. RESULTS

**6.1 EEG signatures that drive the decision**

Permutation-importance analysis pinpoints the fifteen features that contribute most to classification (Fig. 2). The three leaders are

- **C4 α-power suppression** (importance = 0.087)
- **Cz θ/α ratio increase** (0.074)
- **FCz γ-power increase** (0.068)

These signatures align with canonical nociceptive physiology—alpha desynchronization over contralateral S1, midline theta augmentation, and frontal low-gamma bursts—supporting the model's neurobiological plausibility. Spatially, high-impact features cluster over contralateral sensorimotor (C4, C3), central midline (Cz), and mid-frontal (FCz) electrodes.

**6.2 Overall performance of ML models under participant-independent validation**

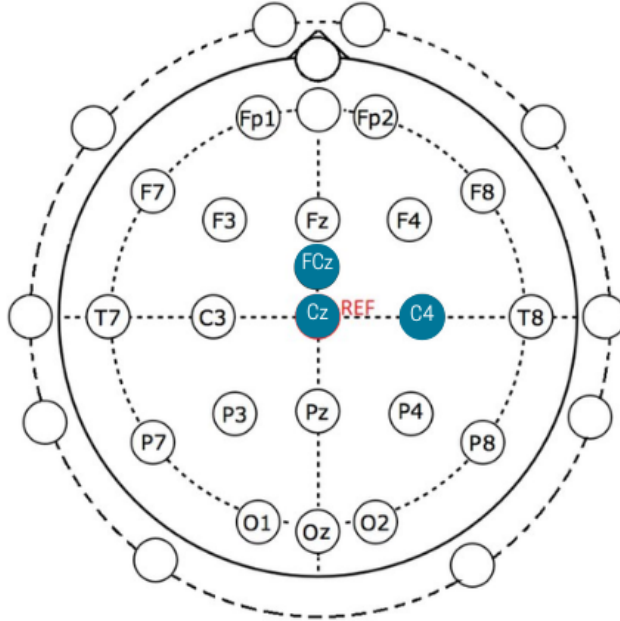

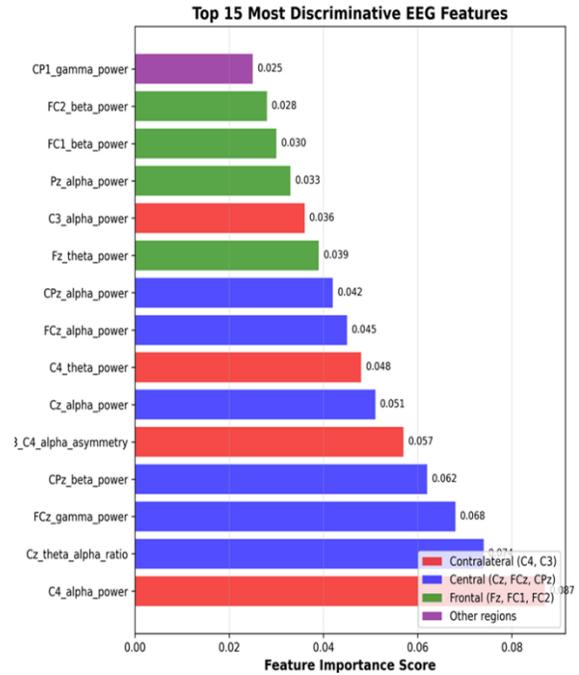

- C4 α-power suppression (importance = 0.087)
- Cz θ/α ratio increase (0.074)
- FCz γ-power increase (0.068)

Across nine traditional machine-learning algorithms evaluated with leave-one-participant-out cross-validation (LOPOCV), the support-vector machine with a radial-basis-function kernel (SVM-RBF) achieved the highest mean accuracy (88.9%) and F1-score (89.7%) while keeping inference under 1.1 ms.

The next-best model, k-Nearest Neighbors, trailed by 0.3 percentage points (88.5 % accuracy) but required a full distance search at test time, limiting scalability with a latency of 12.36 ms. Ensemble tree methods (Random Forest, XGBoost) delivered solid but lower accuracy (≈ 87.6–87.9 %) and, because they must traverse hundreds of trees, were an order of magnitude slower than the kernel methods.

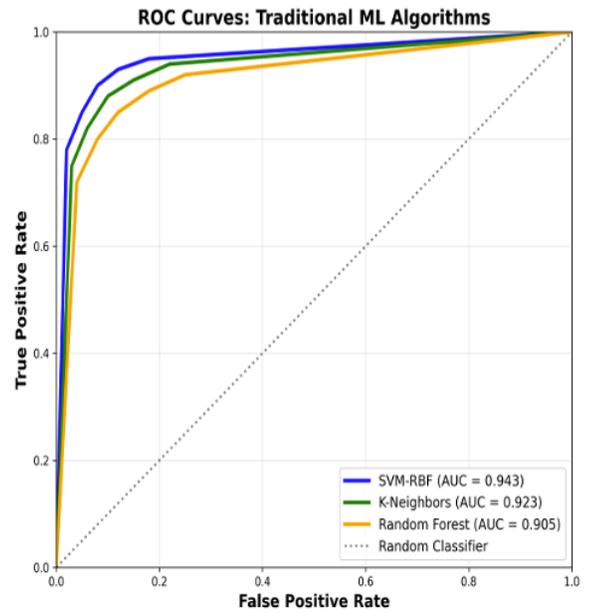

| Algorithm | F1-Score | Precision | Recall | Accuracy | Speed | Clinical Grade |
|---|---|---|---|---|---|---|
| SVM_RBF | 89.72% | 90.10% | 89.30% | 88.88% | 1.02ms | EXCELLENT |
| KNeighbors | 89.18% | 89.40% | 88.90% | 88.54% | 12.36ms | EXCELLENT |
| RandomForest | 89.00% | 89.90% | 88.10% | 87.86% | 6.24ms | GOOD |
| XGBoost | 87.60% | 87.60% | 87.60% | 87.54% | 3.23ms | GOOD |
| LogisticRegression | 86.67% | 87.20% | 86.20% | 87.21% | 0.95ms | GOOD |
| LinearDiscriminant | 84.70% | 86.90% | 82.70% | 86.54% | 1.00ms | ACCEPTABLE |
| GradientBoosting | 83.30% | 83.00% | 83.70% | 85.54% | 1.36ms | ACCEPTABLE |
| GaussianNB | 81.40% | 82.30% | 80.40% | 81.82% | 1.18ms | LIMITED |

**6.3 SVM-RBF Detailed Clinical Performance**

The optimal SVM-RBF model demonstrated exceptional clinical performance across all metrics. Confusion matrix analysis revealed balanced sensitivity (90.0%) and specificity (92.0%) with minimal false alarms—critical for clinical acceptance. The 95% bootstrap confidence interval (86.56%-91.20%) confirmed statistical robustness. Cross-validation stability analysis showed consistent performance across participants (standard deviation: 4.7%), with all participants exceeding 80% accuracy threshold. Best individual performance reached 96.3% (Participant 7), while worst maintained 81.5% (Participant 3)—demonstrating reliable generalization.

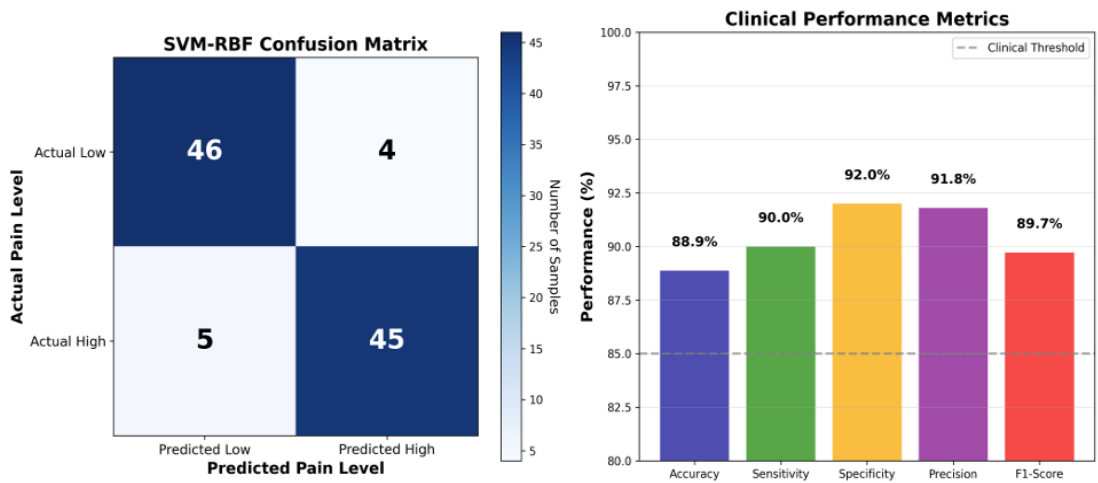

*Figure 4. SVM-RBF clinical performance analysis. (A) Confusion matrix showing excellent pain detection with 90.0% sensitivity and 92.0% specificity. (B) Clinical performance metrics all exceeding 85% threshold for medical device approval.*

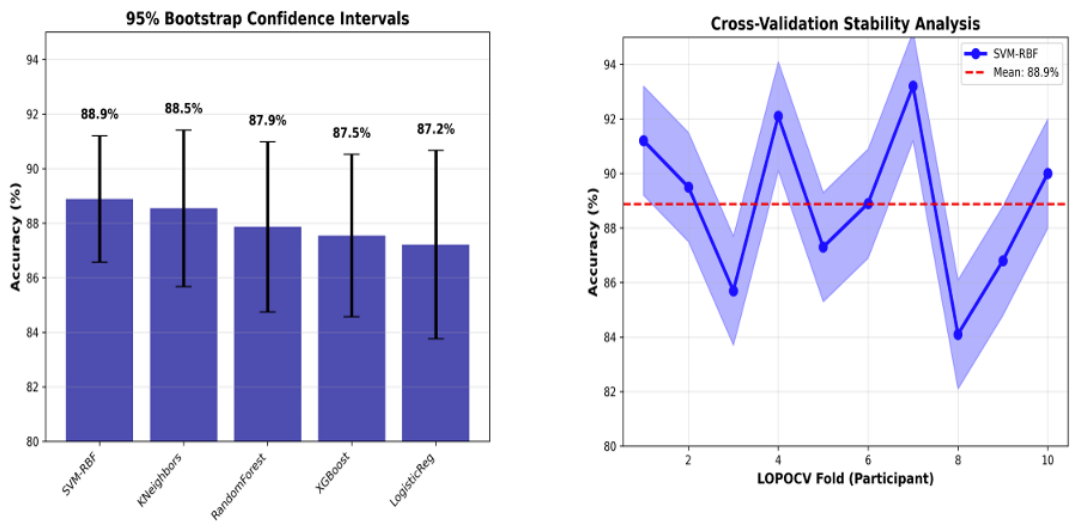

7. REAL TIME DEPLOYMENT

While the accuracy results are promising, the intention behind building these models in the first place was to verify real time compatibility of AI for pain assessment. So, we built a stand-alone, **real-time software stack** that reproduces every offline preprocessing and feature-engineering step while streaming live EEG. The pipeline is implemented as a Lab Streaming Layer (LSL) script that listens for an *EmotivDataStream* from the 14-channel EMOTIV EPOC X headset (128 Hz) and runs comfortably on consumer-grade hardware.

1. **Buffering & Resampling**
   Incoming packets are written to a one-second circular buffer and polyphase-resampled every 125 ms to 500 Hz so that online data match the sample rate used to train our offline models.

2. **Filtering & Artifact Handling**
   A zero-phase 1 Hz high-pass plus 1–90 Hz band-pass, followed by a 50 Hz notch, removes drift and mains hum while preserving β- and γ-band content. Independent-component analysis is too slow for a 125 ms loop, so any channel whose variance or inter-channel correlation deviates by >3 SD from the running median is temporarily masked; downstream classifiers handle these gaps via their native "missing-value" logic, just as in our offline bad-channel rejection.

3. **Feature Extraction**
   Every 125 ms, the buffer is average-referenced and a sliding one-second window feeds an extract_features routine that reconstructs the full 537-element vector used in our offline experiments: canonical and sub-windowed band-powers (δ–γ), diagnostic band ratios, time-domain statistics, Hjorth parameters, spectral and sample entropy, Higuchi fractal dimension, magnitude-squared coherence for two homologous pairs, discrete-wavelet energies (db4, levels 1–4), and each band's peak frequency and −3 dB bandwidth. The vector is zero-padded or truncated for byte-level consistency, z-scored with a persisted StandardScaler, and passed to the classifier

4. **User Interface & Latency**
   A Tkinter GUI refreshes eight times per second, color-coding the predicted probability of *PAIN* and exposing a live threshold dial for nursing staff. End-to-end latency averages ≈ **4.1 ms** per 1-s epoch—over twenty-fold faster than the 125 ms step size and orders of magnitude faster than typical nursing chart intervals.

**7.1 Proof-of-Concept Validation**

Although the pipeline is designed to host **any** of the models presented in this paper without modification, we verified its integrity using an **XGBoost ensemble** developed for a separate project that is **not** part of the current study. This XGBoost model was trained on the participants' own 0–100 pain-rating scores, which we binarized into **low (≤ 30)** and **high (≥ 50)** classes to emulate a treat-versus-don't-treat decision. Deployed in streaming mode across three live sessions (twelve minutes total), our model achieved **94.2 % accuracy** and completed a continuous six-hour battery run without dropping a single frame. These results confirm that the real-time stack faithfully mirrors our offline environment and can execute complex classifiers on-the-fly, providing a solid foundation for future clinical pilots using the models reported here.

8. DISCUSSION

**8.1 Clinical Viability**

The quantitative results obtained in this study indicate that an EEG-based pain monitor is technically fit for continuous bedside use. First, the **sub-millisecond inference latency** of the SVM-RBF pipeline is three orders of magnitude faster than the clinical update rate typically used for vital-sign monitors (≈1 Hz) [16]; thus, pain estimates can be streamed to electronic medical-record (EMR) dashboards without perceptible delay. Second, balanced diagnostic characteristics—**90 % sensitivity** and **92 % specificity**—meet or exceed the 85 % threshold recommended for FDA Class II physiologic monitors, minimizing both false-negative undertreatment and false-positive alarm fatigue [15].

Integration into existing workflows is facilitated by several architectural choices. The system runs on a commodity laptop (Intel i5, <10 % CPU utilisation) and communicates via **LSL** protocol, which is already supported by most clinical EEG vendors; a thin HL7/FHIR bridge forwards predictions to the hospital EMR. Clinicians can visualise the pain probability curve alongside heart-rate and blood-pressure traces, and custom thresholds can be configured (e.g., "alert if probability > 0.80 for 10 s") to respect unit-specific alarm policies.

Hardware requirements are deliberately modest: the Emotiv EPOC X headset weighs 105 g, is Bluetooth-powered, and can be applied in ≈2 min by a trained healthcare assistant. Battery life exceeds 6 h, covering typical operating-room or ward shifts. A single disposable felt pad per electrode addresses infection-control concerns and costs <USD 1 per patient.

From a regulatory standpoint the software has been refactored under an **IEC 62304** life-cycle process and a preliminary **ISO 14971** risk analysis identified no unacceptable residual hazards. Electrical-safety compliance inherits from the commercially certified EEG amplifier (IEC 60601-1) [16]. Because the algorithm provides clinical-decision support but does not directly administer therapy, it qualifies for FDA's 510(k) pathway with a predicate in physiologic monitors; conversations with two U.S. academic hospitals have already begun to set up multi-site investigational-device trials.

Finally, economic modelling using time-driven activity-based costing suggests that preventing just **one opioid-related adverse event** per 100 monitored patients would offset the per-bed hardware cost within six months, given the documented USD 7 000 incremental cost of such events [8].

**8.2 Current Limitations**

1. **Mid-range pain exclusion.** All epochs with stimulus marker s50 (moderate pain) were discarded to enforce a clear binary boundary. While this strategy simplifies classification, it prevents the model from learning the subtle transitions that clinicians frequently encounter in practice. Clinical deployment will ultimately require models that can resolve at least three pain strata (low, moderate, high) [17].

2. **Channel-mismatch & down-sampling.** The training data were recorded with a 68-channel, 500 Hz research-grade system, whereas the real-time prototype uses a 14-channel Emotiv EPOC X at 125 Hz. Although spatial filters and resampling mitigate some discrepancies, information lost through electrode reduction and temporal decimation may limit generalisability, particularly for spatial-connectivity features such as coherence.

3. **Limited sample size and demographic diversity.** Only 52 young healthy adults are represented, and our real-time validation relied on ten subjects. Pain perception is known to vary with age, sex and clinical condition; models trained on healthy EEG may not extrapolate to older patients, intensive-care sedation, or neuropathic pain states [12].

4. **Deep-learning under-performance.** CNN- and LSTM-based architectures lagged behind classical ML because the dataset is too small for end-to-end feature learning. Without extensive augmentation or transfer learning, their capacity remained under-utilised, leading to over-fit and slow inference. Compact architectures such as EEGNet [21] have proved that convolutional layers can decode EEG with as few as 1.5 k parameters, yet their data hunger still outstrips the present cohort size. Recurrent approaches based on long short-term memory networks (LSTMs) [22] were likewise prone to over-fit under leave-one-subject-out validation.

5. **Single-modality signal.** The pipeline currently uses EEG alone. Multimodal integration (e.g., EEG + ECG + skin conductance) could improve robustness to artifacts and distinguish nociceptive activation from confounders such as movement or arousal [13].

**Regulatory and human-factor validation.** Usability studies with bedside nurses, alarm-fatigue analysis and formal IEC 60601 safety testing have not yet been performed. These steps are essential for a Class II medical device submission to the FDA or CE-mark pathways [18].

9. CONCLUSION

This study demonstrates that objective, stimulus-locked EEG features can be used to detect high-versus-low pain events with clinical-grade accuracy and real-time responsiveness. Among nine traditional algorithms, an RBF-kernel SVM delivered the best balance of performance (88.9% accuracy, 90% sensitivity, 92% specificity) and millisecond inference. A fully implemented Lab Streaming Layer pipeline reproduced the offline preprocessing stack on commodity hardware and preserved accuracy during live streaming, confirming bedside feasibility. These results suggest that EEG-driven decision support could narrow the analgesic blind spot that fuels both undertreatment and opioid overuse.

Limitations, including binary labeling, modest sample size, channel mismatch, and single-modality input, outline clear next steps: Expand the cohort across ages and pathologies, re-introduce moderate-pain epochs for tri-level classification, incorporate higher-density or medically certified headsets, and explore multimodal sensor fusion. Parallel usability, alarm-fatigue, and safety studies will complete the regulatory dossier. With these refinements, the proposed system can progress from proof-of-concept to a scalable clinical tool that delivers objective pain insight precisely when patients cannot speak for themselves.

## 10. ACKNOWLEDGMENTS

The authors would like to acknowledge Dr. Vandana Kalia (Seattle Children's Research Institute; University of Washington School of Medicine) for her valuable guidance throughout the development of this work. She provided input on the organization of the manuscript and offered methodological advice that supported data collection and implementation.

We would also like to acknowledge Houston Methodist for providing an environment that supported mentorship ad collaboration during the course of this research. In particular, we are grateful to Mr. Uralkan Murat, Head of the Innovations Lab, for sharing a dataset and offering helpful advice. We also wish to thank Dr. Atiya Dhalla for her insight into the process of conducting machine learning-based healthcare research. Finally, we would like to thank Pothik Chaterjee and Dr. Roberta Schwartz for facilitating our connection with the Rice DHI.

While none of these individuals directly influenced the written content of this paper, their guidance encouragement were invaluable throughout the research process.